\documentclass[10pt,twocolumn,letterpaper]{article}

\usepackage[pagenumbers]{cvpr} 

\usepackage[most]{tcolorbox} 

\usepackage{stfloats} 

\usepackage{float} 

\definecolor{cvprblue}{rgb}{0.21,0.49,0.74}
\usepackage[pagebackref,breaklinks,colorlinks,allcolors=cvprblue]{hyperref}

\newcommand{\name}{\textbf{nanoAstra}}

\def\confName{CVPR}
\def\confYear{2027}

\title{A Glimpse of Training VLMs as Foundation Models for Robotic Manipulation}
\newcommand{\authormail}{\mbox{\tikz[x=1ex,y=1ex,baseline=-0.1ex]{\draw[line width=0.35pt] (0,0) rectangle (1,0.65); \draw[line width=0.35pt] (0,0.65) -- (0.5,0.28) -- (1,0.65);}}}
\author{
Meng-Hao Guo$^{1,*,\dagger}$, Yi Zhang$^{1,*}$, Zhe-Han Mo$^{1,*}$, Han-Zheng Wang$^{1,*}$, Jia-Jun Wang$^{1,*}$,\\
Xin-Sheng Chen$^{1,*}$, Jiaxi Song$^{1}$, Yi-Xuan Deng$^{1}$, Jia-Ning Li$^{1}$, Xuanyu Chu$^{1}$,\\
Hengrui Zhang$^{1}$, Huan-ang Gao$^{1}$, Yongming Rao$^{2}$, Shi-Min Hu$^{1,\authormail}$\\[0.5em]
$^{1}$Tsinghua University \qquad $^{2}$Tencent Hunyuan\\[0.3em]
\url{https://nanoastra.github.io/}\\[0.3em]
{\small $^{*}$Core Contributors \qquad $^{\dagger}$Project Lead \qquad $^{\authormail}$Corresponding author}
}

\begin{document}
\maketitle

\begin{abstract}
Recently, GPT-6 Astra demonstrates strong generalization in robot control, yet limited disclosure of its training leaves open how such capabilities can be reproduced.
We introduce \name{}, an approach to training vision-language models (VLMs) for generalizable manipulation that combines semantic action modeling with capability-preserving supervised fine-tuning (CP-SFT).
\name{} represents robot behavior as semantic steps composed of reusable, parameterized action primitives, guided by explicit progress descriptions.
Meanwhile, CP-SFT uses an explicit robot-specific system prompt to constrain manipulation outputs to a structured action space, aiming to focus adaptation on robot-specific contexts and promote specialization while minimizing interference with broader pretrained capabilities.
With this formulation, we train \name{} on more than 5,000 hours of robot data together with general VQA data.
On RoboTwin 2.0's randomized benchmark, \name{}-9B achieves a 17.8\% zero-shot success rate, substantially outperforming advanced VLA and WAM baselines as well as similarly sized VLMs.
With inference faster than the corresponding motion execution on a single H20 GPU, our model supports action-chunk-like streaming execution that overlaps inference with motion, reducing the robot's idle time from 35.0\% to 6.9\%.\end{abstract}
    
\section{Introduction}
\label{sec:intro}

\begin{figure}[!t]
    \centering
    \includegraphics[width=\linewidth]{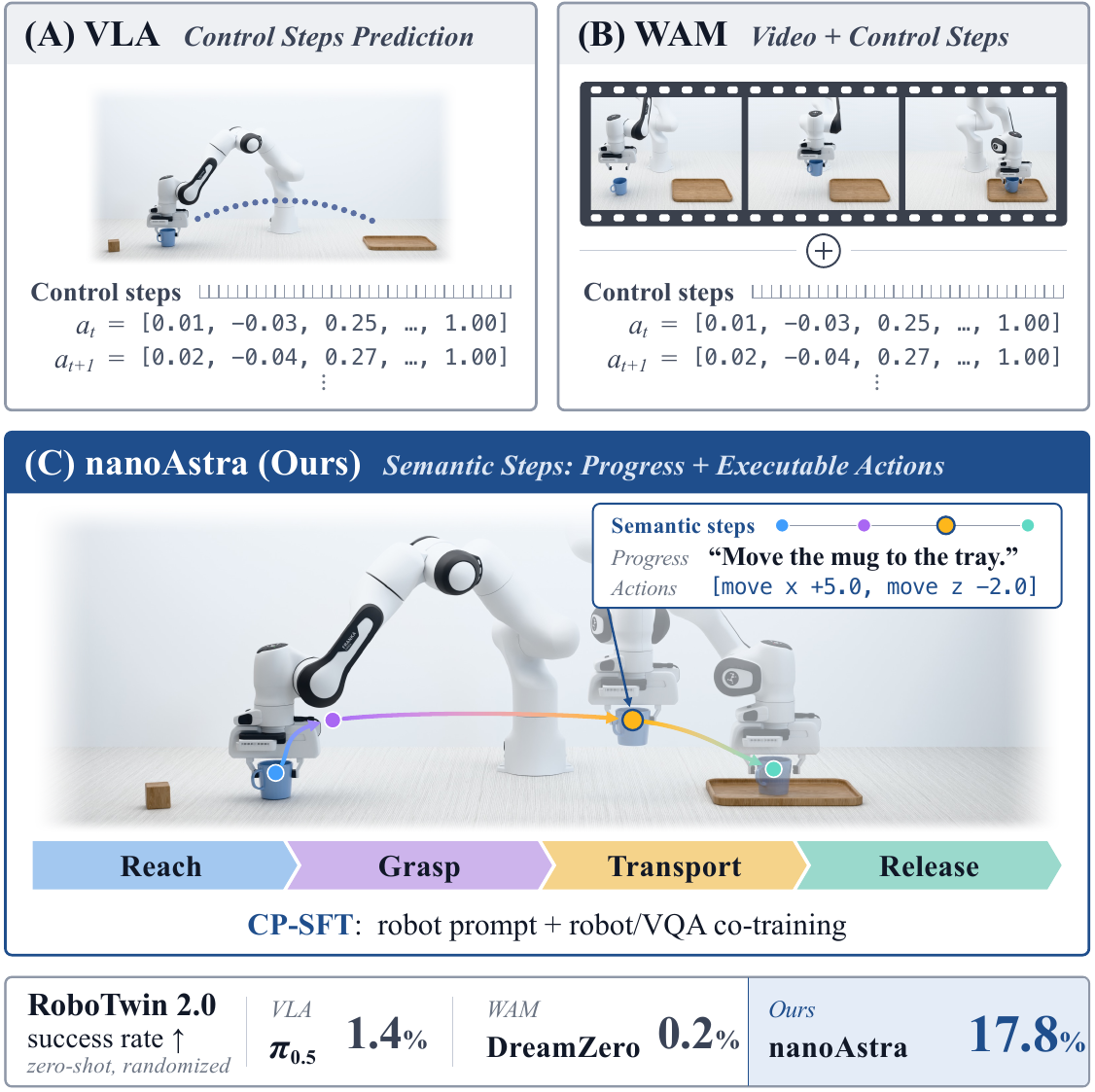}
    \caption{Comparison of VLA, WAM, and \name. Unlike VLA and WAM, which rely on dense control-step prediction, \name{} organizes robot behavior into reusable semantic steps with explicit progress guidance and executable actions.}
    \label{fig:motivation}
\end{figure}

Robot manipulation foundation models aim to acquire generalizable knowledge from diverse experience and apply it to unfamiliar objects, instructions, and environments. 
Recent evaluations of GPT-6 Astra~\cite{astra} suggest that general-purpose VLMs can exhibit substantial manipulation capabilities without task-specific fine-tuning~\cite{zhang2026astraeval}.
However, limited disclosure of Astra's training recipe leaves open how such capabilities can be reproduced through an accessible training approach. This motivates us to investigate how large-scale robot experience can be represented and used to train VLMs toward robot manipulation foundation models.

Current robot manipulation foundation models primarily fall into two categories: vision-language-action (VLA) models and world-action models (WAMs).
VLA models transfer pretrained visual and linguistic knowledge to robot control, 
while WAMs leverage future visual prediction to support action learning~\cite{brohan2023rt2,kim2024openvla,ye2026dreamzero}. 
 Despite substantial progress, generalization to unfamiliar task configurations and novel skill compositions remains challenging~\cite{guo2026vpp2}. 
 We hypothesize that one contributing factor is how robot demonstrations are organized for supervision. Many existing policies~\cite{physicalintelligence2025pi05,ye2026dreamzero}, as shown in \cref{fig:motivation}, predict action chunks~\cite{zhao2023act} composed of densely sampled \emph{control steps}, such as $[(0.01,-0.03, \ldots, 1.00), (0.02,-0.04,\ldots, 1.00), \ldots]$, where each vector specifies motion and gripper commands at a sampled instant.
Although physically well-defined, these targets do not explicitly identify coherent motion units or their roles in the task.
As a result, they lack an easily learnable and reusable structure for representing behaviors. 
The same underlying behavior can yield different sequences depending on control frequency, execution speed, and local corrections. 
Learning from such supervision requires models to recover reusable behavior from fragmented trajectory samples, potentially encouraging dependence on demonstration-specific execution patterns.


In this paper, we present \name{}, an approach to training VLMs for generalizable manipulation through two complementary designs: semantic action modeling and  CP-SFT.
As shown in \cref{fig:motivation}, semantic action modeling organizes robot behavior from densely sampled \emph{control steps} into \emph{semantic steps}, grouping reusable, parameterized action primitives under explicit progress guidance.
Each primitive specifies the acting arm, a motion type, and its geometric parameters, as in \texttt{move x +5.0}, or a gripper operation such as \texttt{gripper open}.
The distinction lies in both the content and granularity of supervision. We extract decision points from motion and gripper transitions, organizing demonstrations into behavior-based steps rather than uniformly sampled control targets. Observations at these decision points are paired with structured responses containing progress descriptions and executable action sequences. This changes what constitutes a training example: the model learns to select and parameterize meaningful motion units in context. We hypothesize that this organization can reduce sensitivity to incidental trajectory details and promote the reuse of familiar primitives across different tasks and environments.
Moreover, high-level task instructions alone often lack sufficient detail about intermediate goals and execution steps, providing limited guidance for determining the next action. To address this, dense progress descriptions specify what the robot should accomplish, while action primitives define how it should execute the task.


The second design aims to equip VLMs with manipulation skills while largely preserving their general VQA capabilities. To this end, we adopt CP-SFT to acquire manipulation skills while minimizing interference with the VLM's pretrained capabilities.
A robot-specific system prompt defines the embodiment, coordinate conventions, and executable action space, framing manipulation as a context-conditioned vision-language task. This conditioning is intended to focus adaptation on manipulation-relevant behaviors while limiting interference with the VLM's broader pretrained capabilities.
Besides, we jointly train on robot demonstrations and general VQA data, providing continued supervision for visual understanding and grounding alongside action learning. 
This design aims to specialize the model's behavior under robotic contexts while mitigating catastrophic forgetting~\cite{mccloskey1989catastrophic,kirkpatrick2017overcoming} of its broader capabilities.



With this formulation, we train \name{} on more than 5,000 hours of robot data together with general VQA data, using the VLM's native text-generation interface without a separate action-prediction head. On RoboTwin 2.0's randomized benchmark, \name{}-9B achieves a 17.8\% zero-shot success rate, 
outperforming advanced VLA and WAM baselines as well as similarly sized VLMs.
On a single H20 GPU, inference is faster than the corresponding motion execution, enabling action-chunk-like streaming execution that overlaps inference with motion to reduce waiting time and improve policy execution efficiency. These findings support \name{} as a promising approach to training VLMs as robot manipulation foundation models.

Our contributions are threefold:
\begin{itemize}
\item We develop a data pipeline that curates and transforms more than 5,000 hours of heterogeneous robot demonstrations into VLM-compatible supervision, organizing behavior into executable semantic steps with dense progress descriptions.
\item We introduce CP-SFT and a joint data recipe as an initial exploration of training VLMs toward robot manipulation foundation models, aiming to acquire manipulation skills while largely preserving pretrained capabilities.
\item We demonstrate strong zero-shot performance on RoboTwin 2.0's randomized benchmark, outperforming advanced VLA and WAM baselines as well as similarly sized VLMs, and improve execution efficiency through streaming inference. These findings provide initial evidence for training VLMs as embodied foundation models and offer a reproducible path toward the generalizable manipulation capabilities observed in Astra.
\end{itemize}

\section{\name{}}

\begin{figure*}[!t]
    \centering
    \includegraphics[width=\textwidth]{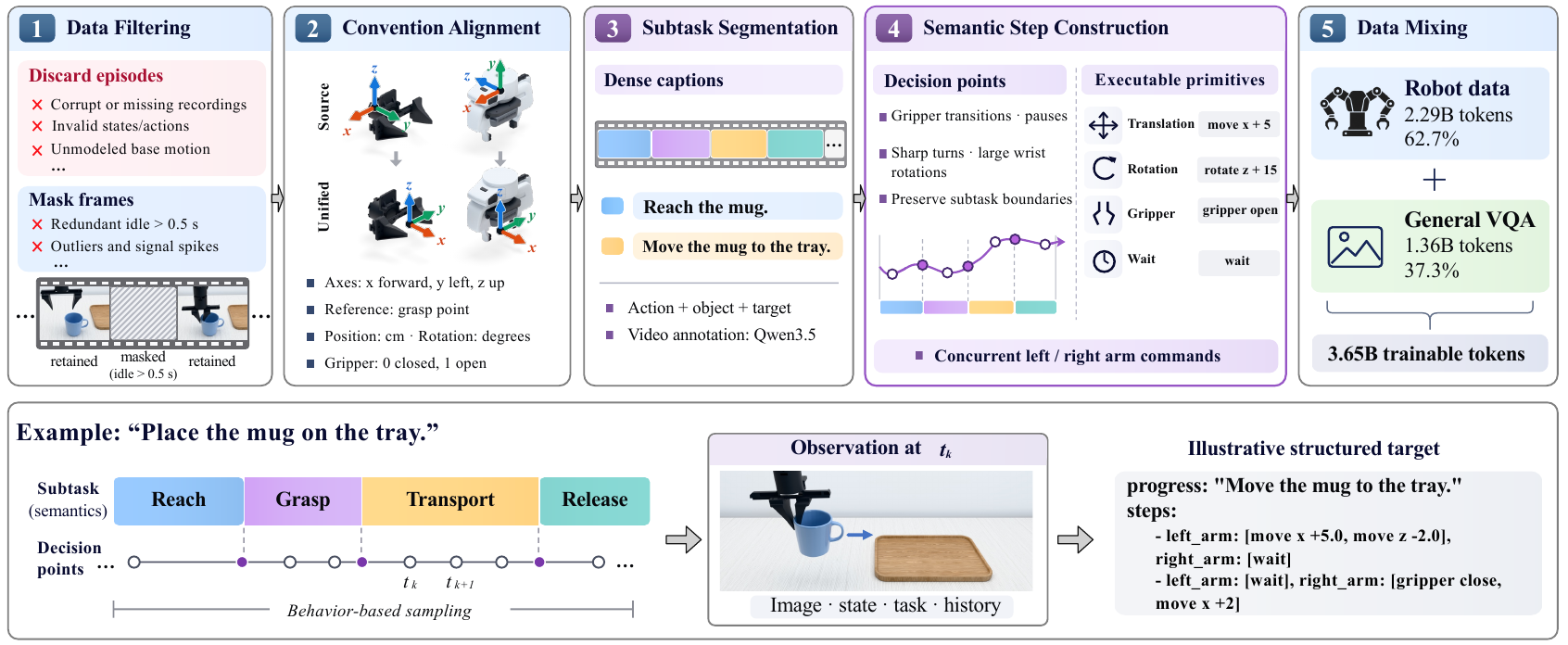}
    \caption{Overview of the \name{} data pipeline. Heterogeneous demonstrations are filtered, geometrically standardized, segmented into subtasks, and converted into semantic steps before mixing with general vision-language data. The example illustrates how observations at behavioral decision points are paired with progress descriptions and executable action primitives.}
    \label{fig:method}
\end{figure*}


\subsection{Data Processing}
\label{sec:data_processing}

We process heterogeneous manipulation data from ABC-130K~\cite{allshire2026abc}, AgiBot~\cite{agibot2025world}, HiFi-UMI~\cite{simpleai2026hifiumi}, and HY-UMI~\cite{tencent2026hyumi}, 
spanning real robot demonstrations and hand-held gripper recordings.
As illustrated in \cref{fig:method}, our pipeline unifies these heterogeneous data into a consistent, VLM-compatible supervision format for semantic action modeling through five stages: data filtering, convention and camera alignment, subtask segmentation with dense captions, semantic step construction at behavioral decision points, and data mixing with general vision-language examples. The resulting unified robot samples pair observations with dense progress descriptions and reusable action primitives.

\begin{figure}[!t]
    \centering
    \includegraphics[width=\linewidth]{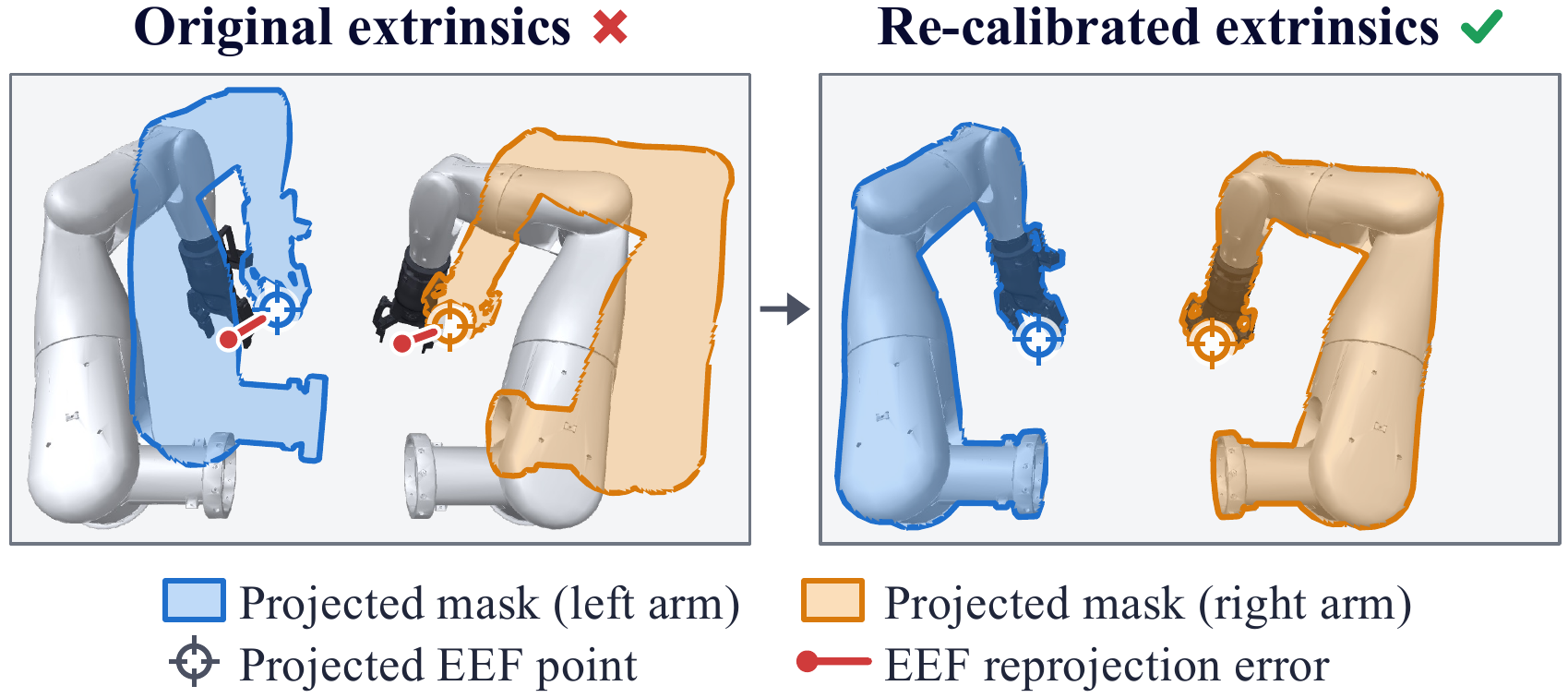}
    \caption{\textbf{Camera re-calibration.}
    We project the robot model at the recorded joint states into the camera image, giving arm masks and end-effector (EEF) points.
    \emph{Left:} with the original extrinsics they are offset from the arms; red segments connect each projected EEF point to the EEF's actual image location (red dot).
    \emph{Right:} after re-calibration they align more closely with the arms, which also reduces the EEF reprojection error.}
    \label{fig:calib}
\end{figure}

\paragraph{Data Filtering.}
We filter demonstrations at both episode and frame levels to obtain reliable supervision for semantic step construction.
At the episode level, we discard recordings with corrupted timing or content, such as abnormal frame rates or durations, timestamp gaps, video--state desynchronization, missing videos, or invalid values~\cite{DBLP:journals/corr/abs-2606-17846}.
We also discard episodes whose states and actions are inconsistent, whose arm or gripper signals are physically implausible, in which the mobile base moves beyond what our arm-only action interface can express, or whose camera extrinsics fail quality checks and cannot be reliably re-calibrated, e.g., when projected arm silhouettes or marker points still deviate substantially from their observed image locations (\cref{fig:calib}).
At the frame level, we mask redundant idle segments before, after, and between motions (pauses longer than 0.5\,s), together with action outliers, jerky motions, abrupt signal spikes~\cite{savitzky1964smoothing,DBLP:journals/corr/abs-2606-17846}, and extreme values that would distort normalization; an episode is dropped only when such frames dominate it.
Filtering at both levels removes unusable recordings while retaining valid portions for constructing temporally coherent action steps.

\paragraph{Convention and Camera Alignment.}
A shared action vocabulary requires consistent geometric meanings for its parameters.
We standardize reference-frame conventions, gripper-axis definitions, and pose representations for end-effector states and action targets across datasets.
Robot demonstrations use a robot-base reference frame, whereas hand-held grippers use a fixed scene frame.
Dataset-specific quaternion layouts and 6D rotation representations~\cite{zhou2019continuity} are converted to a common quaternion representation.
In the model's text interface, positions and translations are expressed in centimetres and rotation angles in degrees; gripper opening is mapped to $[0,1]$, with 0 closed and 1 open.
To reduce dependence on dataset-specific absolute positions, we additionally translate the coordinate origin by an offset sampled independently on each axis within $\pm 30$\,cm and held fixed throughout an episode.

Datasets also differ in their definition of the end-effector (EEF) reference point, which may lie at the wrist, a tracked marker, or the fingertips.
Where kinematic models and gripper geometry allow us to compute the midpoint between the gripper fingertips, we transform the recorded EEF positions to this common grasp reference point; otherwise, we retain the dataset's stated reference point.
The same transformations are applied to current states, history, and action targets, with relative motions computed from the transformed poses.
This standardization gives the action primitives consistent units and explicitly defined reference frames across the training data.

Coordinate conventions must also hold between the text and the image.
Robot states and action targets are defined in the robot-base or scene frame, while the model sees the scene through camera images, and the camera extrinsics link the two.
Some datasets do not release camera extrinsics, and released extrinsics can be inaccurate, in which case EEF positions projected into the image miss the visible grippers.
For these datasets, we re-calibrate the main camera to obtain more accurate extrinsics (\cref{fig:calib}).

\begin{figure*}[t]
    \centering
    \includegraphics[width=\textwidth]{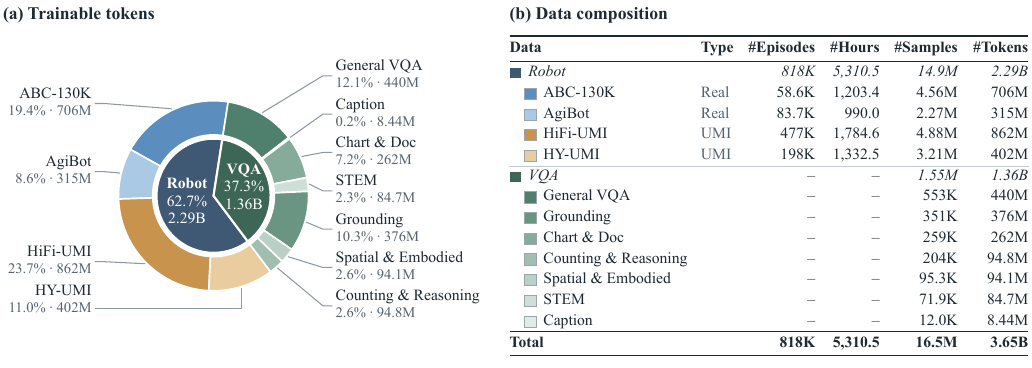}
    \caption{Training data composition for \name{}. (a) Distribution of trainable tokens across robot and general vision-language data. (b) Dataset statistics, including episodes, hours, training samples, and trainable tokens.}
    \label{fig:data_composition}
\end{figure*}

\paragraph{Subtask Segmentation with Dense Captions.}
An episode-level instruction often omits the intermediate goals of a long sequence.
We use gripper transitions and manipulation events to identify temporally localized operation segments, then annotate their associated video clips with Qwen3.5-122B-A10B~\cite{qwen2026qwen35}.
The captions specify the action, manipulated object, and target, yielding a semantic timeline aligned with the demonstration.
These temporally localized captions serve as dense progress descriptions for the subsequent action steps: each step is associated with a local goal that explains its role in the overall task.

\paragraph{Semantic Step Construction.}
Subtasks remain too coarse to serve as action units: grasping an object, for example, can involve reaching toward it and then closing the gripper.
We therefore segment each episode using the unified end-effector trajectories and gripper states, placing cuts at gripper transitions, pauses, sharp turns, and large wrist rotations.
Subtask boundaries are always retained as cuts, so no step spans two subtasks.
Each interval defines a coherent movement and/or gripper operation, and its start frame serves as a decision point for observation and action prediction.
Sampling at these points reduces redundant supervision from nearly identical control frames and organizes targets around changes in behavior.

We express each interval as a semantic step composed of reusable, parameterized action primitives for translation, rotation, gripper control, and waiting.
Commands such as \texttt{move x +5.0} and \texttt{gripper open} explicitly specify motion types and their associated parameters.
Each step contains separate command lists for the left and right arms; commands within an arm's list specify a combined motion, and the two arms execute concurrently.
An inactive arm is represented by \texttt{wait}.
For each training sample, the structured response contains the current progress description in \texttt{progress} and an ordered list of executable semantic steps in \texttt{steps}.
This representation exposes shared motion structure across demonstrations while retaining the geometric arguments required to specify execution targets.

\paragraph{Data Recipe with VLM Data.}
After filtering and processing, we retain 5,310.5 hours of manipulation data across approximately 818K episodes, yielding 14.9M semantic action training samples.
\Cref{fig:data_composition} summarizes this final training corpus: ABC-130K, AgiBot, HiFi-UMI, and HY-UMI contribute 1,203.4, 990.0, 1,784.6, and 1,332.5 hours, respectively.
These statistics describe the data retained for training after our pipeline, rather than the original source collections.
We mix the resulting robot samples with 1.55M general vision-language examples drawn from Honey-Data-V2~\cite{zhang2026bee,zhang2026honeydatav2} in a common conversational format, covering visual question answering, grounding, chart and document understanding, counting and reasoning, spatial and embodied understanding, STEM, and captioning.
The mixture contains approximately 3.65B trainable tokens: 2.29B (62.7\%) from robot data and 1.36B (37.3\%) from general vision-language data.
Robot examples supervise progress descriptions and semantic action sequences, while vision-language examples retain their original instruction-response supervision.
This mixture supports CP-SFT by maintaining general vision-language supervision alongside manipulation learning, with the aim of limiting interference with pretrained capabilities.

\subsection{Capability-Preserving Supervised Fine-Tuning}
\label{sec:sft}

We develop CP-SFT to acquire manipulation skills while aiming to largely 
preserve the pretrained capabilities that may support generalization.
CP-SFT imposes a soft constraint on robot-conditioned outputs 
and supplements cross-entropy with numerical supervision for action prediction.

\begin{figure}[t]
    \centering
    \includegraphics[width=\linewidth]{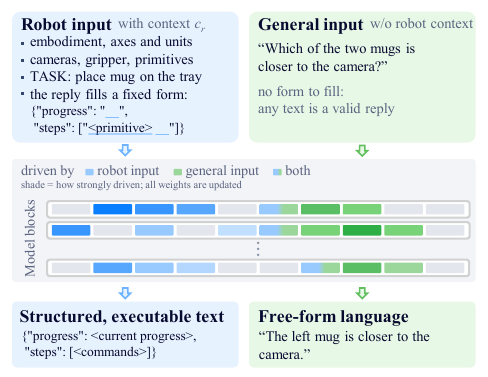}
    \caption{
        \textbf{Softly constrained response decoding space.}
        With the robot system prompt $c_r$, the VLM predicts task progress and executable semantic steps in the structured response space $\mathcal{Y}_r(c_r)$; without $c_r$, the same VLM answers in free-form language.
        The middle panel schematically shows how strongly each input drives the units of every transformer block.
        Together, $c_r$ and the structured targets softly confine manipulation supervision to $\mathcal{Y}_r(c_r)$, aiming to minimize changes to the pretrained output distribution outside robotic contexts.
    }
    \label{fig:cpsft}
\end{figure}

\paragraph{Softly Constrained Response Decoding Space.}
A robot-specific system prompt $c_r$ specifies the embodiment, coordinate conventions, executable action vocabulary, and response format.
Given an input $x$ comprising visual observations, robot states, a task instruction, and interaction history, the VLM predicts a task progress $g$ and executable semantic steps $A$.
We define the structured response space as
\begin{equation}
\mathcal{Y}_r(c_r)=\left\{\operatorname{Serialize}(g,A)\mid g\in\mathcal{G},\ A\in\mathcal{A}(c_r)\right\},
\end{equation}
where $\mathcal{G}$ denotes natural-language progress descriptions and $\mathcal{A}(c_r)$ contains action sequences conforming to the robot interface.
As illustrated in \cref{fig:cpsft}, the system prompt and structured targets jointly impose a soft constraint that confines manipulation supervision to a robot-specific response space. By focusing adaptation within this context-conditioned space, we aim to minimize changes to the VLM’s pretrained output distribution outside robotic contexts, thereby helping preserve its broader capabilities.

\paragraph{Training Objective.}
We augment standard cross-entropy with a modified Wasserstein Number Token Loss~\cite{zausinger2025regress} to improve numerical precision in action prediction.
Our variant introduces decimal-place weighting, explicit sign supervision, and normalization over digit tokens only.
For a number $\nu$, let $D_\nu$ index its digits, where digit $j$ has target value $d_j$ and decimal place $10^{e_j}$ (e.g., $e_j=1,0,-1$ for the three digits of $12.5$).
The place weights are $w_j=10^{e_j}/T_\nu$, with $T_\nu=\sum_{j\in D_\nu}10^{e_j}$.
Let $\widetilde p_j$ be the softmax distribution restricted to the ten digit tokens. We define
\begin{equation}
\begin{aligned}
\ell_\nu^{\mathrm{digit}} &= \sum_{j\in D_\nu}w_j\sum_{v=0}^{9}\widetilde p_j(v)|v-d_j|,\\
\ell_\nu^{\mathrm{sign}} &= \frac{2|y_\nu|}{T_\nu}\sigma(z_{\mathrm{flip}}-z_{\mathrm{true}}),
\end{aligned}
\end{equation}
where $y_\nu$ is the target number and $z_{\mathrm{true}},z_{\mathrm{flip}}$ are logits of the correct sign-bearing token and its sign-flipped counterpart; the sign term is zero without an explicit sign.
Place weighting emphasizes significant digits, the sign term captures errors invisible to digit-only supervision, and digit-only normalization separates numerical proximity from digit probability mass.
Averaging the two terms over numbers $N_s$ in each supervised response gives $\ell_s^{\mathrm{num}}=|N_s|^{-1}\sum_{\nu\in N_s}(\ell_\nu^{\mathrm{digit}}+\ell_\nu^{\mathrm{sign}})$, set to zero when $N_s$ is empty.
The final objective is
\begin{equation}
\mathcal{L}=\frac{1}{|\mathcal{B}|}\sum_{s\in\mathcal{B}}
\left(\ell_s^{\mathrm{CE}}+\lambda\ell_s^{\mathrm{num}}\right),
\label{eq:cp_sft_loss}
\end{equation}
We set $\lambda=0.3$.
Here, $\ell_s^{\mathrm{CE}}$ is the mean cross-entropy over supervised tokens and $\mathcal{B}$ contains all original samples in an optimizer step across GPUs and gradient accumulation, ensuring equal sample weighting regardless of packing.

\section{Experiments}

\subsection{Settings}
We initialize \name{} from Qwen3.5-9B~\cite{qwen2026qwen35} and implement training with ms-swift~\cite{zhao2024swift}.
We train for one epoch on 56 H20 GPUs using the data mixture in \cref{sec:data_processing}.
The effective global batch size is 56 data packs ($\sim$ 188 original samples on average) per optimizer step, with 1 gradient-accumulation step.
We use the AdamW optimizer~\cite{loshchilov2019adamw} with a peak learning rate of $1\times 10^{-5}$, a cosine learning-rate schedule, a warmup ratio of 0.01, and weight decay of 0.1.
The maximum sequence length is 8,192 tokens,
the image-resolution budget is 392 visual tokens per image,   
and the training precision is bfloat16 (mixed precision, DeepSpeed ZeRO-2~\cite{rajbhandari2020zero}).
The trainable parameter scope is full fine-tuning of the entire 9B vision-language model.

\begin{table}[t]
    \centering
    \caption{Performance on RoboTwin 2.0 under the zero-shot randomized setting. SR denotes success rate; higher is better. The best result is \textbf{bold}, and the second-best results are \underline{underlined}.}
    \label{tab:robotwin_results}
    \small
    \setlength{\tabcolsep}{7pt}
    \resizebox{\linewidth}{!}{%
    \begin{tabular}{llc}
        \toprule
        Family & Model & SR (\%) $\uparrow$ \\
        \midrule
        VLA & $\pi_{0.5}$~\cite{physicalintelligence2025pi05} & 1.4 \\
            & Xiaomi-Robotics-1~\cite{xiaomi2026robotics1} & 2.0 \\
            & DM0.5~\cite{dexmal2026dm05} & 2.4 \\
            & LingBot-VLA-2.0~\cite{wu2026lingbotvla2} & \underline{10.8} \\
        \midrule
        WAM & DreamZero~\cite{ye2026dreamzero} & 0.2 \\
            & LingBot-VA~\cite{li2026causal} & 0.0 \\
        \midrule
                 VLM-based & Qwen3.5-9B~\cite{qwen2026qwen35}  & 8.2 \\
                 w/ thinking & Qwen3.8-27B~\cite{qwenteam2026qwen38}  & \underline{10.8} \\
                 & Qwen3.5-35B-A3B~\cite{qwen2026qwen35} & 9.0 \\
                 & Qwen3.5-397B-A17B~\cite{qwen2026qwen35}  & 9.0 \\
        \midrule   
                  VLM-based & Qwen3.5-9B  & 3.2 \\
                  w/o thinking & Qwen3.8-27B  & 3.6 \\
                   & Qwen3.5-35B-A3B  & 3.4 \\
                   & Qwen3.5-397B-A17B  & 4.0 \\
                  & \name-9B  (Ours) & \textbf{17.8} \\
        \bottomrule
    \end{tabular}%
    }
\end{table}

\subsection{Benchmarks}
\paragraph{Zero-Shot Manipulation.}
We evaluate on 50 tasks in RoboTwin 2.0's randomized setting~\cite{chen2025robotwin2}, averaging success rates over 10 runs per task to reduce evaluation variability.
\name{} is trained exclusively on real-world manipulation data, without simulated trajectories, RoboTwin demonstrations, or benchmark-specific fine-tuning; evaluation is therefore zero-shot under the randomized configuration.
The model runs in closed loop: each query, termed a turn, takes the current observation and returns a reply containing the current progress and the semantic steps to execute; an episode ends when the task succeeds or the turn budget is exhausted.

\paragraph{Vision-Language Capability Retention.}
We assess capability retention on MMMU~\cite{yue2024mmmu}, CV-Bench~\cite{tong2024cambrian1}, and ERQA~\cite{deepmind2025erqa}, covering multidisciplinary reasoning, visual perception, and embodied reasoning, respectively.

\subsection{Evaluation on RoboTwin 2.0}
\paragraph{Performance.}
\Cref{tab:robotwin_results} compares VLA, WAM, and VLM-based models under the zero-shot randomized protocol on RoboTwin 2.0.
\name{}-9B achieves the highest success rate of 17.8\%, exceeding both the strongest evaluated VLA, LingBot-VLA-2.0, and the strongest VLM baseline, Qwen3.8-27B with thinking, by 7.0 percentage points.
The evaluated WAMs achieve at most 0.2\%, highlighting the difficulty of transfer under this protocol.

Thinking improves every evaluated Qwen model, raising success rates from 3.2--4.0\% without thinking to 8.2--10.8\% with thinking.
However, thinking generates additional reasoning tokens before action outputs, increasing autoregressive decoding work and potentially delaying action execution in closed-loop control.
Nevertheless, \name{}-9B reaches 17.8\% without thinking, outperforming Qwen3.5-9B by 14.6 points without thinking and 9.6 points with thinking, as well as the larger evaluated Qwen models.
By directly generating structured action responses without an explicit thinking trace, \name{} avoids this additional reasoning-token overhead while achieving higher task success.
These results provide initial evidence that training VLMs with semantic action supervision and CP-SFT can improve zero-shot manipulation beyond the evaluated general-purpose VLM baselines.

\begin{table}[t]
    \centering
    \caption{Inference and execution efficiency of representative VLA, WAM, and VLM-based policies. \textbf{Infer.}: inference latency; \textbf{Motion}: motion execution time; \textbf{I/M}: inference-to-motion time ratio. 
    cmds denotes commands. All inference measurements use a single NVIDIA RTX 4090 GPU, except the result marked with $^{\dagger}$, which uses a single NVIDIA H20 GPU.}
    \label{tab:inference_efficiency}
    \small
    \setlength{\tabcolsep}{4pt}
    \resizebox{\linewidth}{!}{%
    \begin{tabular}{lcccc}
        \toprule
        \textbf{Model} & \textbf{Infer.} & \textbf{Actions} & \textbf{Motion} & \textbf{I/M} \\
        \midrule
        $\pi_{0.5}$ & 101 ms & 45.0 steps & 2.70 s & 0.037 \\
        StarVLA & 70 ms & 16.0 steps & 0.96 s & 0.073 \\
        X-VLA & 143 ms & 28.6 steps & 1.71 s & 0.084 \\
        \midrule
        Fast-WAM & 523 ms & 27.3 steps & 1.64 s & 0.32 \\
        Motus & 1.93 s & 16.0 steps & 0.96 s & 2.01 \\
        LingBot-VA & 8.89 s & 22.2 steps & 1.33 s & 6.67 \\
        \midrule
        Qwen3.5-9B w/ thinking & 60.7 s & 1.6 cmds & 1.42 s & 42.8 \\
        \name{} & 3.89 s & 3.5 cmds & 2.87 s & 1.36 \\
        \name{}$^{\dagger}$ & 1.69 s & 3.5 cmds & 2.66 s & 0.64 \\
        \bottomrule
    \end{tabular}%
    }
\end{table}

\begin{figure}[t]
    \centering
    \includegraphics[width=\linewidth]{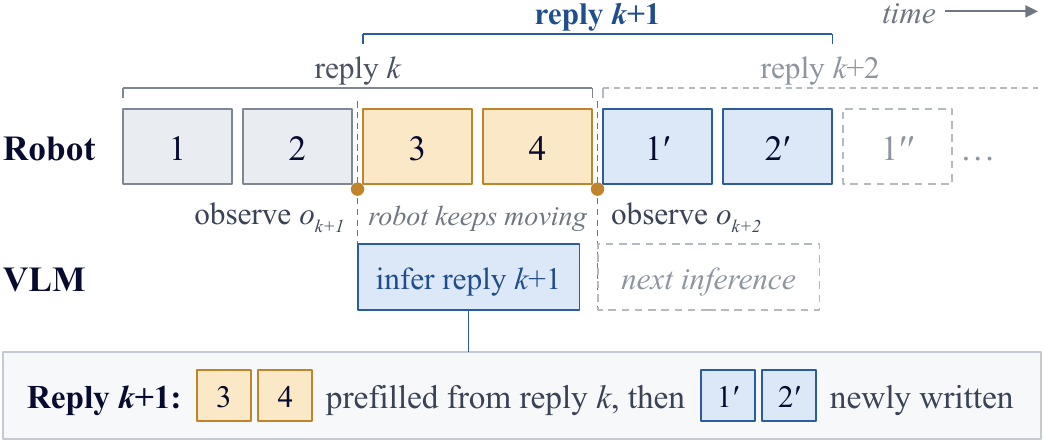}
    \caption{Streaming execution. After step 2 of reply $k$, a new observation is taken and reply $k{+}1$ is requested. The remaining steps (3, 4) are prefilled as its beginning and run during inference; the model then continues with new steps (1$'$, 2$'$).}
    \label{fig:streaming}
\end{figure}

\paragraph{Inference Efficiency.}
\Cref{tab:inference_efficiency} compares inference latency with the duration of the resulting motion.
We define I/M as inference latency divided by the execution time of the motion produced by that prediction; lower values indicate less inference overhead relative to execution.
A ratio above 1 means prediction takes longer than the associated motion, whereas a ratio below 1 indicates potential to overlap inference with execution, provided the system supports pipelining; it does not by itself imply stall-free control.
On a single RTX 4090, \name{} takes 3.89\,s to generate 3.5 commands on average, corresponding to 2.87\,s of motion and an inference-to-motion ratio of 1.36.
Although slower in absolute inference latency than the evaluated VLAs, Fast-WAM, and Motus, it is faster than LingBot-VA (8.89\,s) and has a lower inference-to-motion ratio than both Motus (2.01) and LingBot-VA (6.67).
On a single H20 GPU (marked with $^{\dagger}$), inference latency decreases to 1.69\,s, with 2.66\,s of motion and a ratio of 0.64, making inference shorter than the associated motion execution.

\paragraph{Streaming Inference.}
When execution blocks on inference, the latency of a VLM forces the robot into a stop-and-go pattern: it moves, then waits for the next reply.
On a single H20, inference is already shorter than the motion it produces, so the two can overlap.
We therefore adopt streaming execution (\cref{fig:streaming}): steps are executed as they arrive in the token stream, and the next reply is requested before the current one has finished executing.
At H20 latency, this reduces the share of time the robot spends waiting from 35.0\% to 6.9\% with no loss in success rate (\cref{tab:streaming}).
Inference thus no longer interrupts execution: the robot moves almost continuously through the task rather than pausing at every turn.

\begin{table}[t]
    \centering
    \caption{Blocking vs.\ streaming execution on RoboTwin 2.0. Unlike the standard RoboTwin evaluation, which pauses the simulator during inference, we keep it running as on a real robot, so SR within the official time limit (5 runs per task) is lower than in \cref{tab:robotwin_results}. \textbf{Wait}: wait for inference per turn, in units of one step's duration (0.8\,s on average); \textbf{Idle}: its share of time. Both exclude first turns and turns affected by a failed step.}
    \label{tab:streaming}
    \small
    \setlength{\tabcolsep}{12pt}
    \begin{tabular}{lccc}
        \toprule
        \textbf{Execution} & \textbf{SR (\%)} & \textbf{Wait} $\downarrow$ & \textbf{Idle} $\downarrow$ \\
        \midrule
        Blocking & 11.2 & 2.1 steps & 35.0\% \\
        Streaming & 12.9 & \textbf{0.2 steps} & \textbf{6.9\%} \\
        \bottomrule
    \end{tabular}
\end{table}

\begin{table}[t]
\centering
\caption{Performance on RoboDojo Open. SR denotes success rate (\%); Score denotes the reported progress score. Both are higher-is-better. The best and second-best results are \textbf{bold} and \underline{underlined}, respectively.}
\label{tab:robodojo_results}
\small
\setlength{\tabcolsep}{4pt}
\renewcommand{\arraystretch}{1.15}
\resizebox{\linewidth}{!}{%
\begin{tabular}{llcc}
\toprule
Family & Model & SR (\%) $\uparrow$ & Score $\uparrow$ \\
\midrule
VLA & Xiaomi-Robotics-1 & $0.25$ & $0.25$ \\
& LingBot-VLA 2.0 & 0.00 & 0.00 \\
& $\pi_{0.5}$ & 0.00 & 0.00 \\
& DM0.5 & 0.00 & 0.00 \\
\midrule
WAM & DreamZero & 0.00 & 0.00 \\
& LingBot-VA & 0.00 & 0.00 \\
\midrule
VLM-based & Qwen3.5-9B & 0.00 & 0.00 \\
w/ thinking & Qwen3.5-35B-A3B & $\underline{0.75}$ & $\underline{0.75}$ \\
\midrule
VLM-based & Qwen3.5-9B & $0.25$ & $0.25$ \\
w/o thinking & Qwen3.5-35B-A3B & $\underline{0.75}$ & $\underline{0.75}$ \\
& \name{}-9B (Ours) & $\mathbf{4.75}$ & $\mathbf{5.06}$ \\
\bottomrule
\end{tabular}%
}
\end{table}

\begin{figure*}[t]
    \centering
    \includegraphics[width=\textwidth]{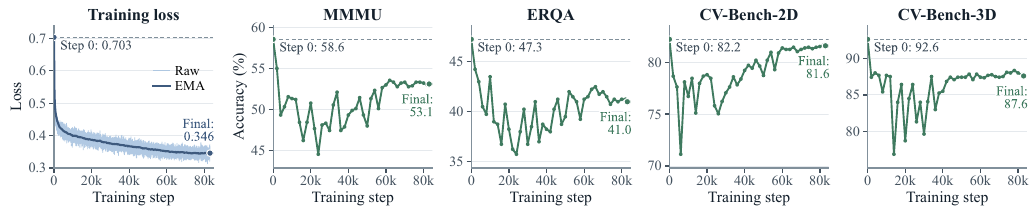}
    \caption{Training loss and benchmark accuracy over training. Dashed lines mark the accuracy at step 0.}
    \label{fig:training_curves}
\end{figure*}

\begin{figure*}[t]
    \centering
    \includegraphics[width=0.95\textwidth]{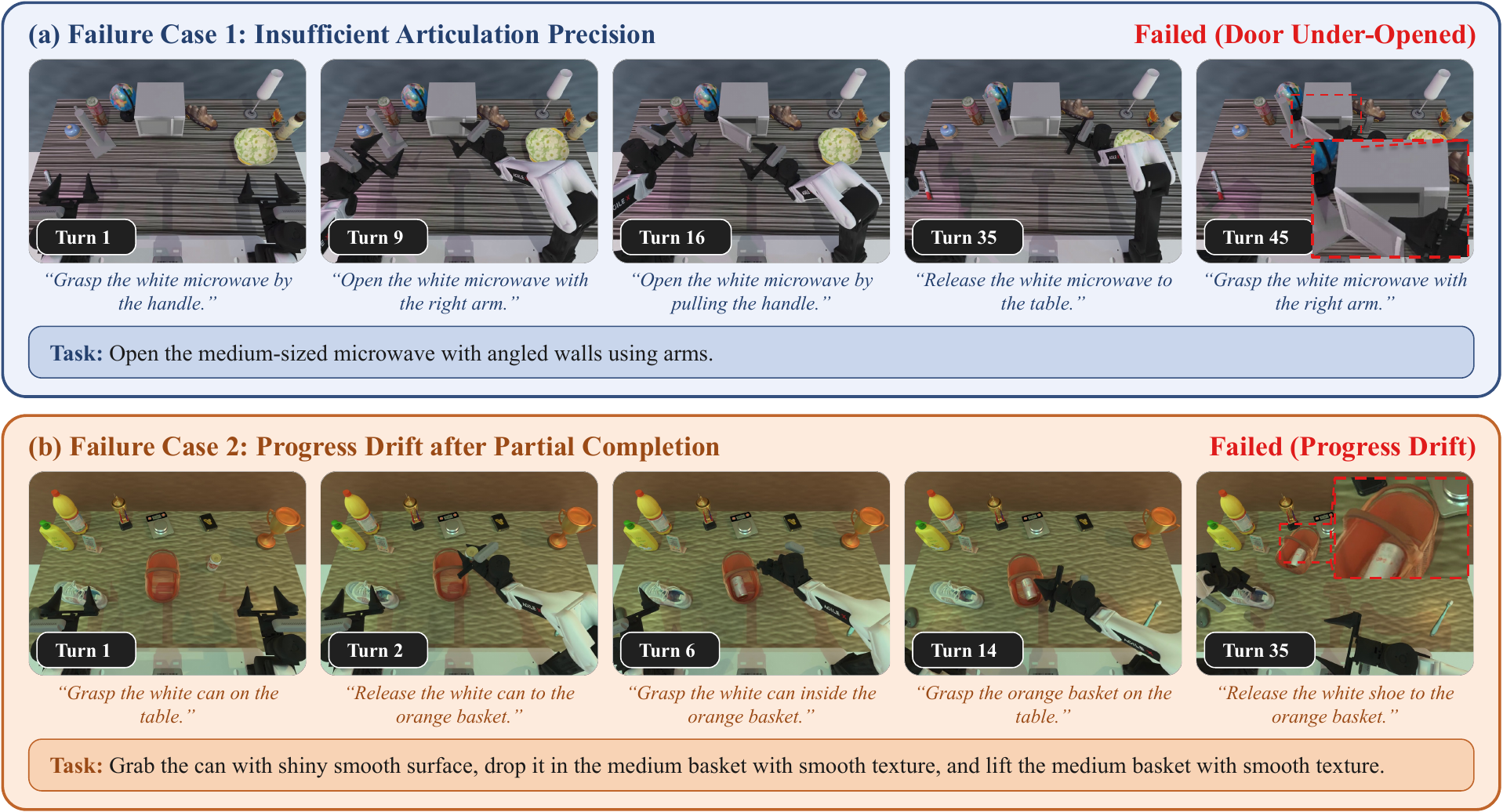}
    \caption{Failure cases of \name{} on RoboTwin 2.0. Each frame shows the decision turn and the predicted progress. 
    }
    \label{fig:failure_cases}
\end{figure*}

\subsection{Evaluation on RoboDojo}
\label{sec:robodojo}
We report additional results on RoboDojo Open~\cite{chen2026robodojo} in \cref{tab:robodojo_results}, following the model-family grouping in \cref{tab:robotwin_results}. The evaluation covers 8 tasks, with 10 seeds and 5 runs per seed, for 400 episodes per model.

\name{}-9B achieves a 4.75\% success rate (19/400), compared with 0.25\% for its untuned Qwen3.5-9B counterpart without thinking. It achieves the highest success rate and progress score among the models reported in \cref{tab:robodojo_results}. These results suggest transfer beyond RoboTwin 2.0.

\subsection{Training Process Monitoring}
\Cref{fig:training_curves} tracks optimization and vision-language capability retention throughout training.
The training loss drops rapidly during early adaptation and then decreases steadily, with its exponential moving average reaching approximately 0.35 by the end of training.
Meanwhile, VQA accuracy initially declines and fluctuates, followed by recovery and greater stability at later checkpoints rather than continued deterioration.

At the final checkpoint, MMMU, ERQA, CV-Bench-2D, and CV-Bench-3D reach 53.1\%, 41.0\%, 81.6\%, and 87.6\%, respectively, compared with initial scores of 58.6\%, 47.3\%, 82.2\%, and 92.6\%.
CV-Bench-2D nearly recovers its initial performance, with a gap of only 0.6 percentage points, while the other benchmarks retain most of their initial accuracy with gaps of 5.0--6.3 points.
These trends indicate that improved fitting of the training data coexists with substantial, though incomplete, retention of broader vision-language capabilities.
We believe that broader and more diverse VQA co-training data could further reduce the degradation while supporting manipulation learning.

\subsection{Failure Analysis}

\Cref{fig:failure_cases} presents two representative failure cases, both of which make substantial progress before exhausting the turn budget. In \cref{fig:failure_cases}(a), the predicted progress largely follows the grasp--open--release cycle of the microwave door. However, each opening step is executed as straight-line translations, and the door opens only partially rather than reaching the success threshold.

In \cref{fig:failure_cases}(b), the model places the can in the basket early (Turn~2), but its progress then returns to the completed subtask (Turn~6). Although it later moves on to grasping the basket (Turn~14), its progress drifts again to a distractor shoe (Turn~35), so the basket is never lifted. These cases indicate that errors in motion parameterization and task-state estimation can accumulate during closed-loop execution. We hypothesize that object-centric primitives for articulated objects and explicit tracking of completed subtasks could further improve long-horizon manipulation.

\section{Related Work}

\subsection{Foundation Models for Robotic Manipulation}

VLA models transfer pretrained visual and linguistic knowledge to robot control, as exemplified by RT-2~\cite{brohan2023rt2}, OpenVLA~\cite{kim2024openvla}, $\pi_{0.5}$~\cite{physicalintelligence2025pi05}, X-VLA~\cite{zheng2026xvla}, and StarVLA~\cite{starvla2026starvla}.
However, action prediction alone does not ensure semantic alignment across heterogeneous data, even when policies perform well within their training domains.
WAMs learn physical dynamics through joint video and action prediction, as exemplified by DreamZero~\cite{ye2026dreamzero}, LingBot-VA~\cite{li2026causal}, Motus~\cite{bi2026motus}, and Fast-WAM~\cite{yuan2026fastwam}.
However, predicted visual dynamics remain tied to embodiment, even when video-based learning improves transfer.
Recently, VLM-based agents, such as RoboDawn~\cite{guo2026robodawn} and Robo-Harness K1~\cite{li2026roboharness}, leverage pretrained VLMs~\cite{astra, fable52} for manipulation, showing early signs of generalization across embodiments, tasks, and environments.
These advances motivate exploring how manipulation capabilities can be acquired through direct VLM training.



\subsection{Data and Data Pipelines for Embodied Models}

Large-scale datasets expand the diversity of embodied experience, including robot collections such as Open X-Embodiment~\cite{openx2023}, DROID~\cite{khazatsky2024droid}, AgiBot~\cite{agibot2025world}, and ABC-130K~\cite{allshire2026abc}.
UMI~\cite{chi2024umi}, HiFi-UMI~\cite{simpleai2026hifiumi}, and HY-UMI~\cite{tencent2026hyumi} further broaden data collection through hand-held manipulation interfaces.
However, scale and diversity alone do not resolve cross-source differences in embodiment, motion representation, temporal granularity, and task semantics.

Data processing pipelines organize these experiences into supervision suited to different model families.
VPP2~\cite{guo2026vpp2} uses detailed captions and event-level manipulation videos to support video pretraining for WAMs.
For VLA training, Qwen-RobotManip~\cite{DBLP:journals/corr/abs-2606-17846} harmonizes heterogeneous datasets through signal filtering, state--action consistency checks, and coordinate alignment.
Our pipeline organizes heterogeneous robot trajectories into executable semantic steps with dense progress guidance, providing structured supervision for VLM training.
We further mix robot samples with general VQA data to support manipulation learning while aiming to preserve pretrained capabilities.

\section{Conclusion}

We present \name{}, an initial exploration of training VLMs as foundation models for robotic manipulation. Combining semantic action primitives with CP-SFT, \name{} learns from over 5,000 hours of robot experience and general vision-language supervision. It achieves promising zero-shot manipulation while largely retaining pretrained capabilities, despite some degradation. Fast inference enables streaming execution that overlaps inference with motion, improving policy execution efficiency. Together, these findings support structured action semantics and capability-aware adaptation as a practical path toward generalizable and efficient manipulation foundation models.


{
    \small
    \bibliographystyle{ieeenat_fullname}
    \bibliography{main}
}


\end{document}